\documentclass[conference]{IEEEtran}
\IEEEoverridecommandlockouts
\usepackage{cite}
\usepackage{amsmath,amssymb,amsfonts}
\usepackage{algpseudocode}
\usepackage{algorithm}
\usepackage{graphicx}
\usepackage{subcaption}
\usepackage{textcomp}
\usepackage{xcolor}

\def\BibTeX{{\rm B\kern-.05em{\sc i\kern-.025em b}\kern-.08em
    T\kern-.1667em\lower.7ex\hbox{E}\kern-.125emX}}
\begin{document}

\title{Control of a Nature-inspired Scorpion using Reinforcement Learning\\
\thanks{$^{\dagger}$ Equal Contribution}%
\thanks{A. Agrawal, V S Rajashekhar and R. Arasanipalai are Research Assistants, and D. Ghose is a professor, at the Guidance, Control, and Decision Systems Laboratory (GCDSL), Dept. of Aerospace, Indian Institute of Science, Bangalore, India.  Emails:juhi05aakritiagrawal@gmail.com; vsrajashekhar@gmail.com; rohitkumar97@gmail.com; dghose@iisc.ac.in}}

\author{\IEEEauthorblockN{Aakriti Agrawal$^{\dagger}$}
\and
\IEEEauthorblockN{V S Rajashekhar$^{\dagger}$}
\and
\IEEEauthorblockN{Rohitkumar Arasanipalai}
\and
\IEEEauthorblockN{Debasish Ghose}
}

\maketitle

\begin{abstract}
A terrestrial robot that can maneuver rough terrain and scout places is very useful in mapping out unknown areas. It can also be used explore dangerous areas in place of humans. A terrestrial robot modeled after a scorpion will be able to traverse undetected and can be used for surveillance purposes. Therefore, this paper proposes modelling of a scorpion inspired robot and a reinforcement learning (RL) based controller for navigation. The robot scorpion uses serial four bar mechanisms for the legs movements. It also has an active tail and a movable claw. The controller is trained to navigate the robot scorpion to the target waypoint. The simulation results demonstrate efficient navigation of the robot scorpion.
\end{abstract}


\section{Introduction}

Robotic scorpions can be used for exploring various planetary terrains hazardous for humans \cite{scoting_leg_rob}. The tail of a scorpion can adjust to balance out the scorpion when walking on uneven terrain. By using a combination of various tail angles, the scorpion robot and cross uneven and steep terrain with minimal effort. A robotic scorpion using a central pattern generator (CPG) has been done in \cite{scor_cont}. In the works of \cite{leg_loss}, the stability of the scorpion robot when it loses the legs are studied. In the works of \cite{rob_scor}, various geometric parameter of a scorpion robot were controlled and accelerated running were observed in the simulation.


A reinforcement learning (RL) setup consists of 2 components, an agent and en environment. The agent is the robot scorpion which interacts with the environment by performing the actions given by the RL controller. The environment is acted on by the agent and after the action is performed, it returns the new state and rewards. Most RL algorithms follow a similar pattern: Firstly, the environment passes the initial state to the agent, which then performs some actions in order to proceed to the next state. The environment then returns the new state along with the reward for the performed action. Based on the reward, the agents learn which action-state pairs maximize the rewards. This loop continues until the terminal state is reached.

We use a model-free RL algorithm with a deterministic policy, as we do not intend to learn the complex dynamics of the environment. This is unlike model-based RL algorithms which need to learn the complete state transition probabilities from the pair of current state and action to the next state.

The organization of the paper is as follows: Section \ref{section:model} describes the URDF and CAD model of the scorpion and Section \ref{section:rl_control} describes the RL based scorpion controller. A thorough comparison of results is done in Section \ref{section:result} and conclusions are drawn in Section \ref{section:conc}.

\section{Model of the Scorpion}\label{section:model}

\begin{figure*}
    \centering
    \begin{subfigure}[t]{0.5\textwidth}
        \centering
        \includegraphics[width=0.5\linewidth,trim={13cm 7cm 25cm 11cm},clip]{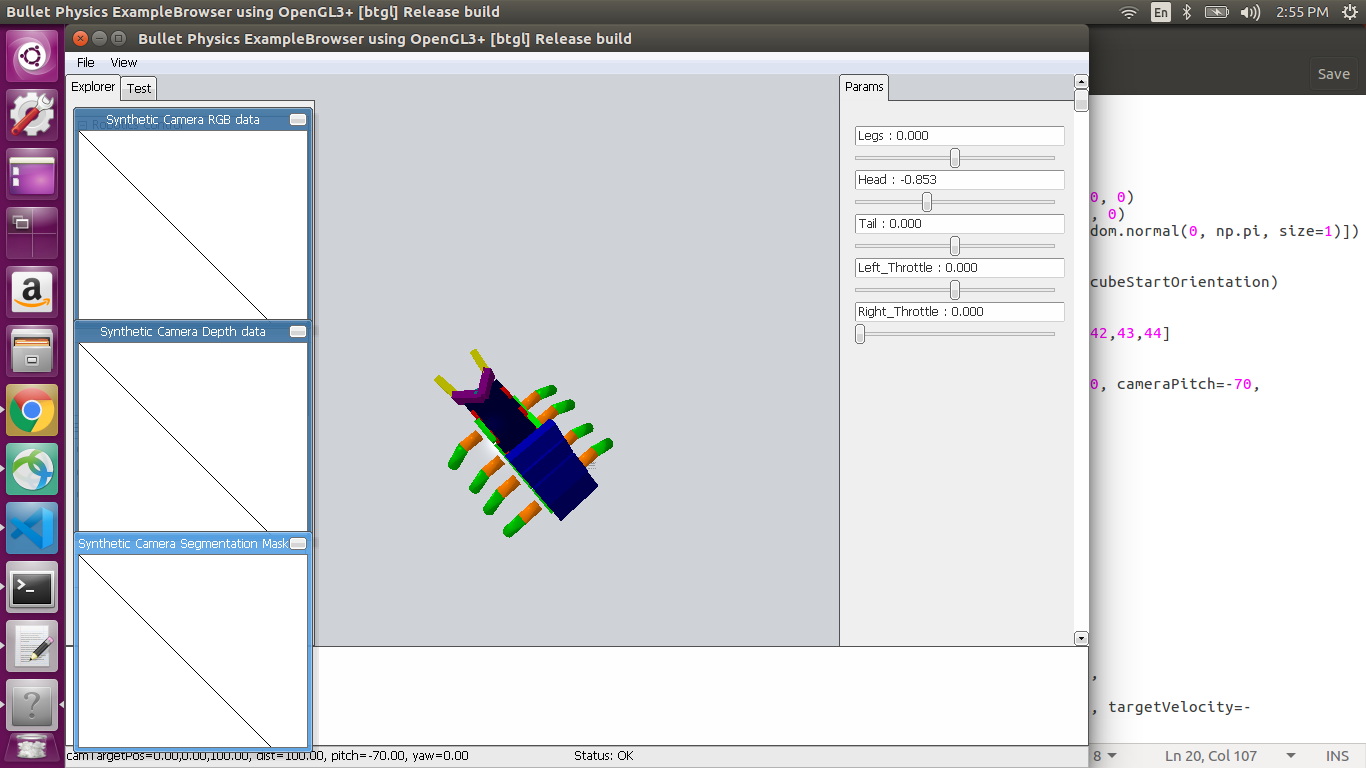}
        \caption{URDF model of the scorpion used for simulation}
        \label{fig:urdf_model}
    \end{subfigure}%
    ~ 
    \begin{subfigure}[t]{0.5\textwidth}
        \centering
        \includegraphics[width=0.35\linewidth,trim={0cm 0cm 0cm 0cm},clip]{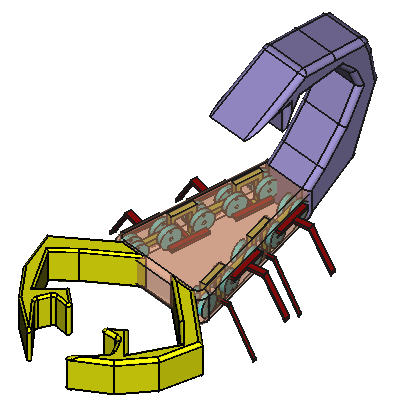}
        \caption{CAD model of the scorpion for experiments}
        \label{fig:cad_model}
    \end{subfigure}
    \caption{Figure showing both the URDF and CAD model of the scorpion}
\end{figure*}

Figure \ref{fig:urdf_model} shows the proposed robotic scorpion that uses two sets serial four-bar mechanism, one on each side to which the legs are attached, for walking. This is controlled using an encoded DC motor. It has an active tail and claw controlled using servo motors. The scorpion was modelled in Unified Robot Description Format (URDF) for simulation. Figure \ref{fig:cad_model} shows the CAD model of the scorpion which is fabricated using a 3D printer. Both models are made to mimic a real scorpion.

\subsection{Simulation Environment}
The simulation environment was created using the python wrapper of the Bullet Real-Time Physics Simulator (PyBullet) \cite{pybullet}. The URDF of the scorpion was imported into pybullet. To make the simulation environment RL algorithm friendly, it was wrapped using OpenAI Gym. This makes the simulation environment RL compatible by ensuring the environment takes as input, the actions and returns, as output, the next state and the appropriate reward.

\section{Controlling the Scorpion}\label{section:rl_control}
This work uses proximal policy optimization (PPO), a model-free RL aglorithm,  in order to control the scorpion. A trained RL network takes as input the current state of the agent, and outputs the most optimal action to be taken by the agent.

\subsection{Policy Optimization}
PPO is a policy optimization algorithm. The general policy optimization algorithm is given in Algorithm \ref{po} where value function network is $V(s|\eta)$ with parameters $\eta$ and our policy is $\pi(s|\theta)$ with parameters $\theta$. $v_{t}$ is defined as
\begin{equation}
    v_{t} = \sum_{i=t}^{T-1} \gamma^{i-1}r_{i} + \gamma^{T-t}V(s_{T}|\eta) \label{eq:1}
\end{equation}
where, $\eta$ are the parameters of the approximated value function, $T$ is the length of the trajectory, $\gamma$ is the discount factor and $r$ is reward as given in (\ref{eq:reward}) below. 

\begin{algorithm}
\caption{Policy Optimization}\label{po}
\begin{algorithmic}[1]
    \State Initialize parameters for $V(s|\eta)$ and $\pi(s|\theta)$
    \State \textbf{while} $j$ = 1,2,3.. until convergence \textbf{do}
    \State\hspace{\algorithmicindent} Collect data according to \textbf{Exploration Strategy}
    \State\hspace{\algorithmicindent} Compute MC estimate of $v^{p}_{t}$ using (\ref{eq:1}) 
    \State\hspace{\algorithmicindent} Update $V(s|\eta)$ $n_{v}$ times.
    \State\hspace{\algorithmicindent} Update $\pi(s|\theta)$ once using standard gradient descent.
    \State \textbf{end while}
\end{algorithmic}
\end{algorithm}

\subsection{State}
The state of the RL agent is position $(x,y)$ and orientation $(\phi, \theta, \psi)$ in Euler angle format, that is, roll, pitch and yaw.
Other state defining parameters like linear velocity and acceleration are not used so as to reduce the dependence of the model on too many sensor data. Let the state be defined by parameter $s$, where $s$ is,

\begin{equation}
s = \{ x, y, \phi, \theta, \psi \}
\label{eq:state}
\end{equation}

We have trained our agent with reasonable environmental constraints in position and orientation. These constraints are, the position of the scorpion is constrained to a $[200,200]$ arena with origin as the centre, and the orientation of the scorpion is always horizontal. The constraint in position helped in normalising the observation space between -1 to 1. This also gives us the benefit of scaling the space during testing even though the model is not trained for an expanded space.

\begin{align*}
x, y &\in [-100,100]\\
z &= 15\\
\phi, \theta, \psi &\in [-\pi,\pi]
\end{align*}

\subsection{Actions}
Our actions comprises of three outputs, left motor speed, right motor speed and tail angle. We believe tail plays an important role in balancing the scorpion and so we include it as an action in the RL model. The actions are also normalised between -1 to 1. 

\begin{equation}
M_{L}, M_{R}, T \in [ -1,1 ]
\end{equation}
where $M_{L}$, $M_{R}$ and $T$ are left motor speed, right motor speed and tail angle respectively. The outputs of the RL agent are scaled appropriately before sending it to the simulation environment. $M_{L}$ and $M_{R}$ are scaled by 3.5 rad/s. As motor speeds above 3.5 rad/s make the scorpion highly unstable and therefore is not recommended. Similarly, $T$ is scaled with 0.365 rad and shifted by a mean of -0.105 rad to scale and constrain the tail angle in the range of $[0.26, 0.47]$ rad. This also puts constraints in the actions which aids in learning the optimum policy.

\begin{figure}
  \centering
  \includegraphics[width=\linewidth,trim={0cm 8cm 2cm 2.5cm},clip]{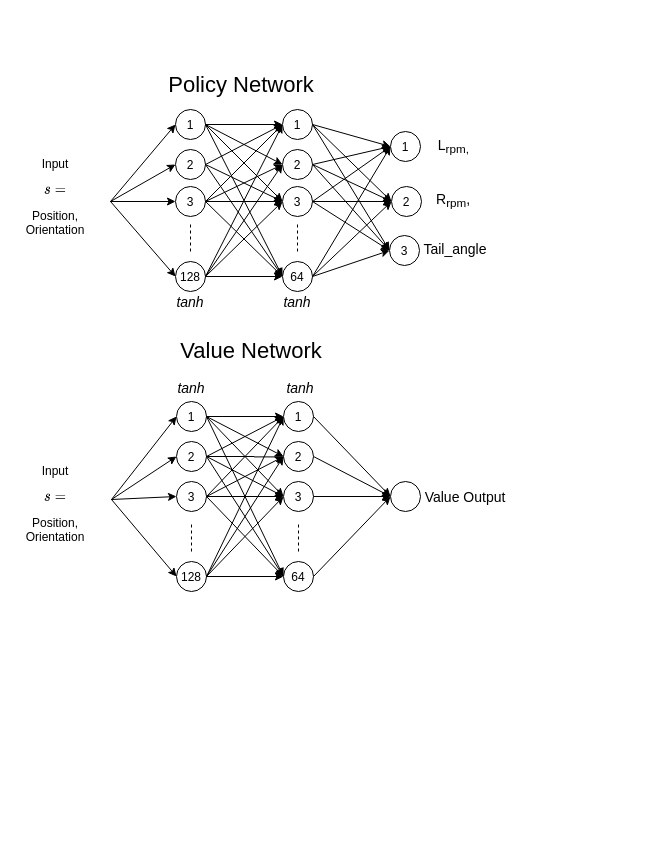}
\caption{The policy network and the value function network.}
\label{fig:network}
\end{figure}

\subsection{Rewards}\label{Reward}
Our aim is to do way-point tracking so the reward is based on position. The reward should increase as the distance to the target point waypoint decreases and vice versa. Therefore, we have used the negative of the euclidean distance in the reward formulation.

\begin{equation}
r_{t} = -2 \times 10^{-3} ||x,y|| \label{eq:reward}
\end{equation}

We take the euclidean distance from the center $(0,0)$ and train our network to track the origin as the target waypoint. While executing position control for some other arbitrary point we can shift the global frame of reference such that the arbitrary point becomes $(0,0)$ point. 

\subsection{Proximal Policy Optimization}
As mentioned, we have used a policy gradient method, Proximal Policy Optimization (PPO) \cite{PPO} with deterministic policy. There exist several policy optimization algorithms alongside PPO like DDPG (Deep Deterministic Policy Gradient) \cite{DDPG} and TRPO (Trust Region Policy Optimization) \cite{TRPO}. Among these, DDPG, has convergence issues and TRPO is complex and computationally intensive. Hence, we selected PPO which has been found to be simpler and more computationally efficient.

The agents consists of two networks for training, a value network, and a policy network. Both networks have 2 hidden layers of 128 and 64 nodes respectively with tanh activation function. The network architecture is given in \ref{fig:network}.

\subsection{Training}
Each trajectory is of 50 seconds with 0.1 sec simulation step. Thus, each trajectory is 500 timesteps long. It takes around 6 hours for the model to train completely and loss to saturate on Nvidia GTX 1080 Ti. The other training parameters are listed in Table \ref{tab:train_param}.

\begin{table}
  \centering
  \resizebox{0.75\columnwidth}{!}{%
    \begin{tabular}{|l|l|}
        \hline
        & \text{Parameter Values}\\
        \hline
        \text{Learning Rate} & $10^{-3}$ \\
        \text{Discount Factor} & 0.99\\
        \text{Entropy Coefficient} & 0.0005\\
        \text{No. of Training Epochs} & 1500 \\
        \hline
    \end{tabular}
    }
    \caption{Training parameters}
        \label{tab:train_param}
\end{table}

\section{Results}\label{section:result}
We evaluated the RL controller in simulation and performed waypoint tracking. We plotted the scorpion position $(x,y)$ along with target waypoint $(x_0,y_0)$ for a single simulation run. We also shift the target waypoint and show that the RL controller is able to do waypoint tracking. We also plot the trajectories of 20 simulations to calculate the failure rate of the controller. Each simulation is run for 1000 timesteps to give it enough time to reach the waypoint.

\subsection{Single simulation Results}
\begin{figure}[t!]
\begin{minipage}{0.5\textwidth}
\begin{subfigure}{\textwidth}
  \centering
  \includegraphics[width=\linewidth]{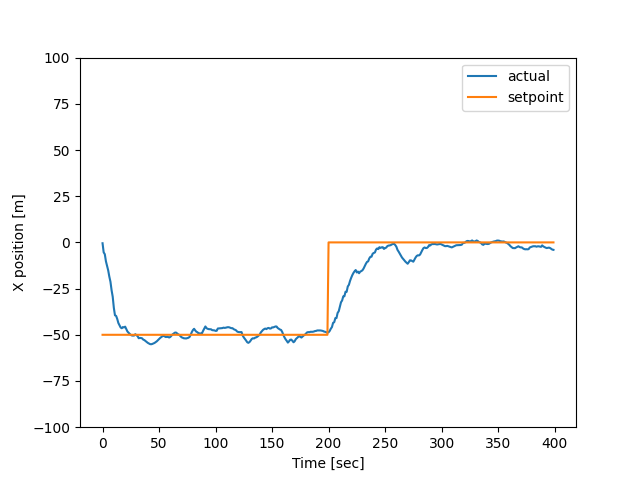}
\end{subfigure}
\begin{subfigure}{\textwidth}
  \centering
  \includegraphics[width=\linewidth,trim={0cm 0cm 0cm 1.cm},clip]{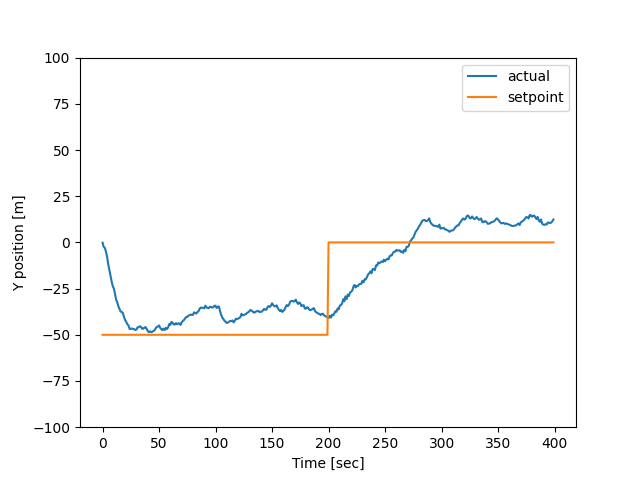}
\end{subfigure}
\vspace{-0.6\baselineskip}
\caption{Figure showing the $x$ and $y$ plots while the scorpion is doing waypoint tracking}
\label{fig:waypoint_tracking}
\end{minipage}
\end{figure}
Figure \ref{fig:waypoint_tracking} shows waypoint tracking of the scorpion RL controller. The graphs show the position $(x,y)$ of the scorpion along with the point it is tracking. The initial target point is $(-50,-50)$ and after 200 seconds the target waypoint is shifted to $(0,0)$. As can be seen from the figure, the scorpion is able to tracking the changing waypoint quite effectively.


\begin{figure}[t!]
\begin{minipage}{0.5\textwidth}
\begin{subfigure}{\textwidth}
  \centering
  \includegraphics[width=\linewidth]{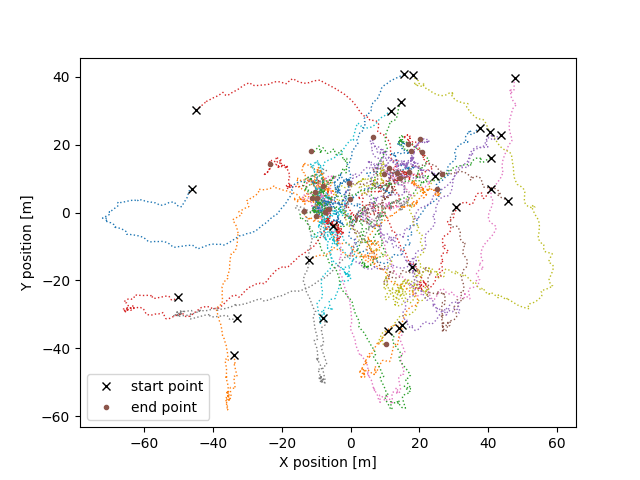}
\end{subfigure}
\begin{subfigure}{\textwidth}
  \centering
  \includegraphics[width=\linewidth,trim={0cm 0cm 0cm 1.cm},clip]{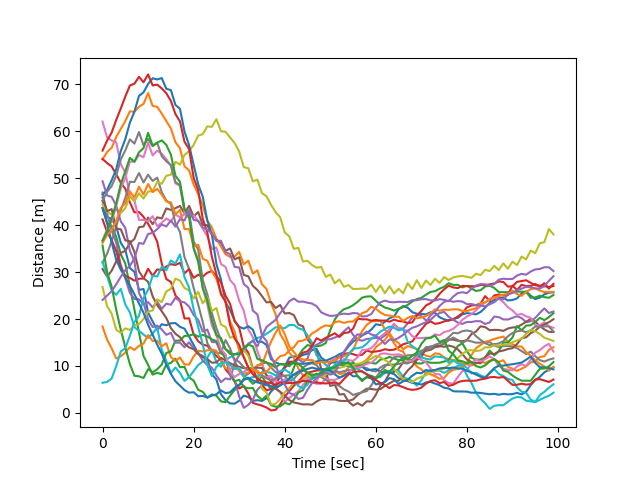}
\end{subfigure}
\vspace{-0.6\baselineskip}
\caption{Figure showing the trajectories and distance from target waypoint for 25 runs of the simulation}
\label{fig:failure_rate_traj}
\end{minipage}
\end{figure}

\begin{figure}[t!]
\begin{minipage}{0.5\textwidth}
\begin{subfigure}{\textwidth}
  \centering
  \includegraphics[width=\linewidth]{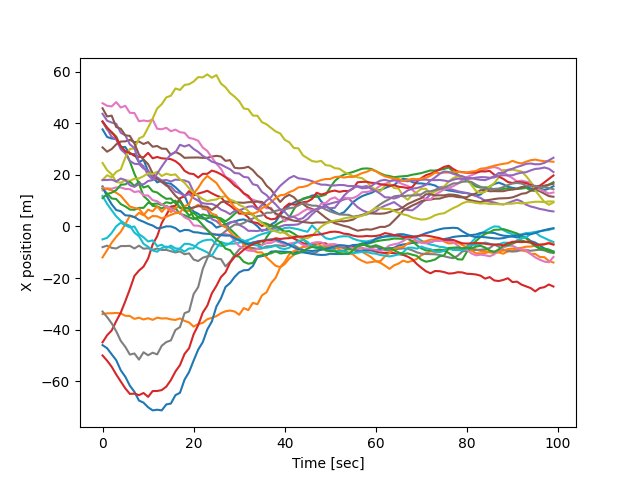}
\end{subfigure}
\begin{subfigure}{\textwidth}
  \centering
  \includegraphics[width=\linewidth,trim={0cm 0cm 0cm 1.cm},clip]{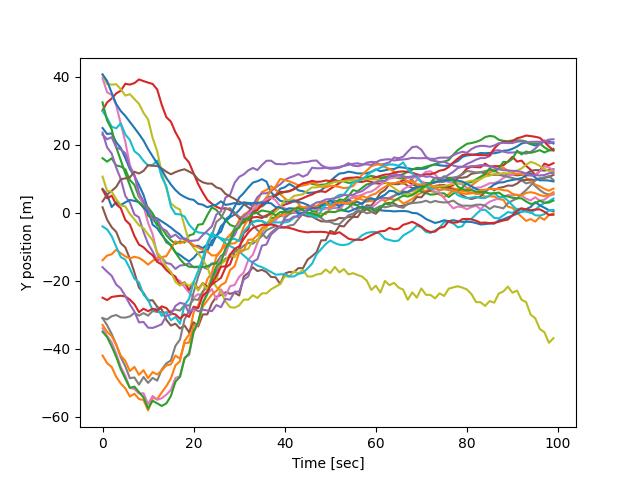}
\end{subfigure}
\vspace{-0.6\baselineskip}
\caption{Figure showing the x and y position of the scorpion for 25 runs of the simulation}
\label{fig:failure_rate_x_y}
\end{minipage}
\end{figure}

\subsection{Failure rate Calculation}

Figure \ref{fig:failure_rate_traj} shows the various trajectories followed by 25 scorpions tracking the origin as the target waypoint. The scorpions were initialized with random position and orientation and their convergence is plotted. As can be seen from the figure, out of 25 scorpions, all but one converged to the target waypoint within the error margin. The average error of the convergence from the tragte waypoint is ... . For further reference, the $x$ and $y$ position of the 25 robot scorpions have been plotted in Figure \ref{fig:failure_rate_x_y}. As can be seen from these plots, the $x$ and $y$ position of the scorpion converges to the target waypoint of $(0,0)$.

\section{Conclusion and Future Work}\label{section:conc}
The main application of this work is in surveillance. Therefore, we plan to implement control of the scorpion even after damage or failure of legs, motors etc. We also aim to train it for various different terrain and elevation height. 

Another future work is to enhance the grabbing mechanism of the scorpion for pick and place tasks. Recurrent neural networks like LSTMs instead of multi-layer perceptron can also significantly improve the performance and thus can be implemented as future work.

\bibliographystyle{IEEEtran}
\bibliography{IEEEexample}

\end{document}